\documentclass{article}

\usepackage[final]{corl_2024} 

\usepackage[utf8]{inputenc} 
\usepackage[T1]{fontenc}    
\usepackage{hyperref}       
\usepackage{url}            
\usepackage{booktabs}       
\usepackage{amsfonts}       
\usepackage{nicefrac}       
\usepackage{microtype}      
\usepackage{xcolor}         
\usepackage{multirow}
\usepackage{multicol}
\usepackage{graphicx} 
\usepackage{amssymb}
\usepackage{xcolor}
\usepackage{amsmath}

\usepackage{hyperref}
\hypersetup{
    colorlinks=true,
    linkcolor=blue,
    filecolor=magenta,      
    urlcolor=cyan,
    pdftitle={Transferable Tactile Transformers},
    pdfpagemode=FullScreen,
    }
\newcommand{\changes}[1]{#1}

\title{Transferable Tactile Transformers for Representation Learning Across Diverse Sensors and Tasks}

\author{
  Jialiang Zhao$^1$ \quad Yuxiang Ma$^2$ \quad Lirui Wang$^2$ \quad Edward H. Adelson$^1$\\
  MIT CSAIL\\
  $^1$\texttt{\{alanzhao,adelson\}@csail.mit.edu}\\
  $^2$\texttt{\{yxma20,liruiw\}@mit.edu}\\
}

\begin{document}

\maketitle

\vspace{-1em}

\begin{abstract}
This paper presents \texttt{T3}: Transferable Tactile Transformers, a framework for tactile representation learning that scales across multi-sensors and multi-tasks.
\texttt{T3} is designed to overcome the contemporary issue that camera-based tactile sensing is extremely \textit{heterogeneous}, i.e. sensors are built into different form factors, and existing datasets were collected for disparate tasks.
\texttt{T3} captures the shared latent information across different sensor-task pairings by constructing a shared trunk transformer with sensor-specific encoders and task-specific decoders.
The pre-training of \texttt{T3} utilizes a novel Foundation Tactile (\texttt{FoTa}) dataset, which is aggregated from several open-sourced datasets and it contains over 3 million data points gathered from 13 sensors and 11 tasks.
\texttt{FoTa} is the largest and most diverse dataset in tactile sensing to date and it is made publicly available in a unified format. 
Across various sensors and tasks, experiments show that \texttt{T3} pre-trained with \texttt{FoTa} achieved zero-shot transferability in certain sensor-task pairings, can be further fine-tuned with small amounts of domain-specific data, and its performance scales with bigger network sizes.
\texttt{T3} is also effective as a tactile encoder for long horizon contact-rich manipulation.
Results from sub-millimeter multi-pin electronics insertion tasks show that \texttt{T3} achieved a task success rate 25\% higher than that of policies trained with tactile encoders trained from scratch, or 53\% higher than without tactile sensing. 
Data, code, and model checkpoints are open-sourced at \url{https://t3.alanz.info}.
\end{abstract}

\section{Introduction}
\label{sec:intro}

The tactile sensing modality has gained increasing popularity within the robotics community, by providing important fine-grained contact information for dexterous and contact-rich manipulation tasks, such as robot grasping \cite{calandra2018more, calandra2017feeling} and fabric manipulation \cite{sunil2023visuotactile, yuan2018active}.
Camera-based tactile sensing \cite{liu2022gelsight}, a sensing method that operates by embedding a camera beneath a soft elastomer to capture the fine-grained interactions with the environment, is among the most popular methods of tactile sensing for its higher resolution and lower cost.
However, camera-based tactile sensors are extremely \textit{heterogeneous}, and there has not been a converged sensor design widely adopted by the robotics community.
Different tactile sensors can differ significantly in shapes, types and numbers of cameras, placement and colors of illumination, etc.
Such inherent heterogeneity hinders roboticists from building a general-purpose tactile encoder that is transferable across different sensors and downstream tasks.

Existing learning architectures and datasets focus on one specific sensor-task pairing, and when it comes to a newly emerged sensor or task, data recollection and training an encoder from scratch are often required.
This issue harms learning efficiency in a more significant way in longer horizon tactile manipulation tasks, where the training of the tactile encoder is guided by a more sparse reward. 
Intuitively, although different tactile sensors produce vastly different tactile images and various tasks extract different information from a tactile input, there should be shareable latent information due to the inherent similarities in tactile sensing across sensors and tasks. 
Therefore, it is both technically viable and practically desirable to design an architecture capable of extracting such shared latent representations and transferring them across different sensor-task pairings.

Learning from heterogeneous tactile inputs shares many similarities with multi-modal representation learning.
Recent works in multi-modal foundation models have demonstrated the potential capability to bridge the gap between different modalities and learning a common latent representation, such as in the perception domain \cite{girdhar2023imagebind, bachmann2022multimae, huang2022amae} as well as the robotic manipulation domain \cite{wang2024poco, shah2023mutex}.
However, learning such foundation models often requires temporally aligned multi-modal inputs or a distance function that describes the resemblance between inputs from different modalities.
In \textit{heterogeneous tactile learning}, the hardware-level disparity often makes it ill-defined to align images collected from different embodiments.

In this work, we tackle the problem of heterogeneous tactile learning using unaligned tactile data, with the goal of learning a scalable representation that can be shared across different sensors and tasks.
We first aggregate existing open-sourced tactile datasets and assemble the Foundation Tactile (\texttt{FoTa}) dataset, which is the largest and most diverse tactile dataset so far to the best of the authors' knowledge.
We then propose Transferable Tactile Transformers (\texttt{T3}) to learn a shared representation across multi-sensors and multi-tasks.
\texttt{T3} is constructed with sensor-specific encoders, a shared trunk, and task-specific decoders.
Our experiments show that \texttt{T3} pre-trained with \texttt{FoTa} achieves reasonable zero-shot transferability, it can be further fine-tuned with small domain-specific datasets, and its performance scales with bigger network sizes.
We also demonstrate that \texttt{T3} can be used as a tactile encoder for longer-horizon manipulation policies with 3 behavior cloning-based sub-millimeter level electronics insertion tasks.
In summary, this paper proposes:

\paragraph{Foundation Tactile (\texttt{FoTa}):}
a dataset containing $3,083,452$ tactile images collected with 13 sensors for 11 tasks in a unified format.
It is aggregated with several of the largest open-sourced tactile datasets and additional data collected in-house by the authors. 

\paragraph{Transferable Tactile Transformers (\texttt{T3}):}
a neural network framework that learns a shared representation across all sensors and tasks in \texttt{FoTa}. Pre-trained weights are provided, and we demonstrate that \texttt{T3} can be easily fine-tuned to a new task or a new sensor with few fine-tuning data. 

\begin{figure}[!t]
\centering
\includegraphics[width=\linewidth]{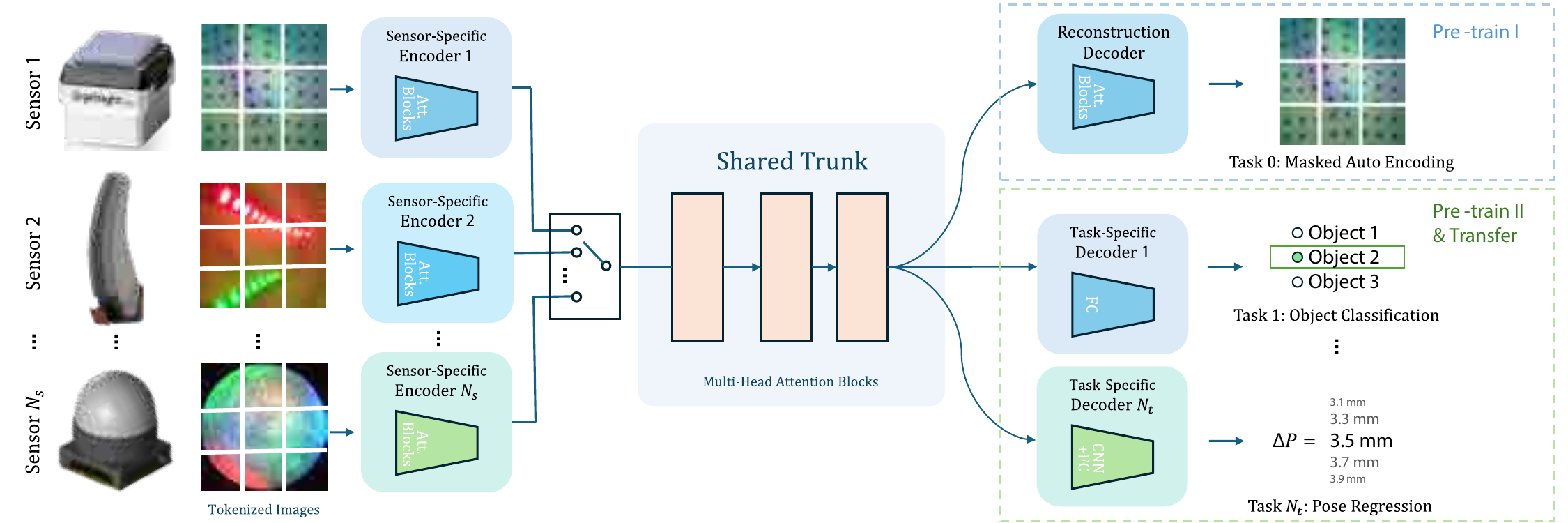}
\caption{\textbf{Architecture illustration of Transferable Tactile Transformers (\texttt{T3}).}
\texttt{T3} learns a shared representation across heterogeneous tactile sensors and downstream tasks with a shared trunk between sensor-specific encoders and task-specific decoders.
The encoders and the shared trunk are constructed with transformer blocks.
The decoder architectures are chosen according to the types of the tasks: we use transformer for generative tasks like reconstruction for masked auto encoding, MLP for classification tasks, and CNN + MLP for pose estimation.
}
\label{fig:net}
\end{figure}

\section{Related Works}
\label{sec:related_works}

\paragraph{Camera-based tactile sensors}


Camera-based tactile sensors are often constructed with a soft elastomer as the contact medium, a reflective layer, a light source to provide illumination, and one or more than one camera to capture the deformation of the elastomer, which therefore provides a detailed view of the contact surface.
Specific design choices in camera-based tactile sensors are highly diverse, such as monochrome~\cite{si2024difftactile} v.s. colored~\cite{gelsight2017, wang2021gelsight, do2022densetact} in illumination, with v.s. without markers~\cite{li2018slip, yuan2015measurement, yamaguchi2016combining}, and different form factors such as flat~\cite{gelsight2017, wang2021gelsight}, dome-shaped~\cite{tippur2023gelsight360, do2022densetact}, finger-shaped~ \cite{zhao2023gelsight, sun2022soft}, compliant ~\cite{liu2023gelsight,ma2024scalable}, multi-linked~ \cite{ma2024gellink} or even palm-shaped~\cite{zhao2023novel}.
Such diversity and disparity have enabled researchers to use those sensors across a spectrum of applications such as in defect inspection \cite{jiang2021vision} and robotic tool-using \cite{wang2024poco}.
Nevertheless, they also add a substantial barrier in leveraging machine learning to encode the information coming from those sensors.
A unique sensor-task pairing often requires data re-collection and neural network re-training.

\paragraph{Datasets for tactile manipulation}
Several large-scale tactile sensing datasets have been released in recent years tackling visual-tactile cross-modality reasoning~\cite{Li_2019_CVPR, fu2024tvl, suresh2021efficient}, object or material classification~\cite{suresh2021efficient, yang2022touch, yuan2018active}, longer horizon tasks such as robotic grasping stability analysis~\cite{calandra2017feeling}, or shape reconstruction~\cite{gao2022objectfolder}.
Those datasets also cover interactions with a wide variety of objects, including household objects and toys \cite{calandra2017feeling, Li_2019_CVPR, fu2024tvl, suresh2021efficient, gao2022objectfolder}, fabrics and clothes \cite{yuan2018active}, natural objects such as wood and gravel \cite{yang2022touch}.
Quantity-wise, several of those datasets contain well over a million images\cite{Li_2019_CVPR, gao2022objectfolder}, making the total number of available tactile images on par with ImageNet-1K \cite{deng2009imagenet}.

While quantity is often not an issue when using existing open-sourced tactile datasets, one significant drawback is that all the aforementioned datasets lack diversity in terms of sensors and tasks.
The vast majority of tactile datasets were collected entirely with only one specific tactile sensor. 
With this constraint, even if a sufficiently well-performing tactile encoder can be trained from one dataset, such a pre-trained encoder will not be able to transfer to another sensor.
In order to learn a general-purpose representation for tactile sensing like \cite{he2016deep,radford2021learning} does in the natural image domain, a large and diverse tactile dataset with a wide variety of sensor-task pairings is a prerequisite.

\paragraph{Representation learning with heterogeneous data}

Learning from heterogeneous data domains, such as taking multi-sensory feedback as inputs or predicting outputs for different tasks, has been an active area of research, especially in the robot learning domain. 
Many existing works combine tactile, vision, and occasionally other modalities such as robot states and sound, by passing each modality through one modality-specific encoder, then concatenating them and passing them through a decoder network to predict the final output \cite{li2022see, li2018slip, qi2023general, wang2024poco, wang2024hpt}.
Apart from learning-driven neural networks, it is also common to use analytical tools, such as factor graphs or convex optimization, to combine predictions from multiple domains~\cite{zhao2023fingerslam}.

Another branch of multi-modal robot learning actively reasons the relationship among heterogeneous domains and cross-links them to learn a shareable latent representation.
Extensive research has explored using contrastive learning to join temporally aligned modalities~\cite{girdhar2023imagebind, shah2023mutex, dave2024multimodal}. 
To connect vision and touch, Chen et al. designed a multi-modal self-attention mechanism~\cite{chen2022visuo} to learn a shared representation, while Li et al. proposed a cross-modal Generative Adversarial Network and were able to predict tactile images from visual images and vice versa \cite{Li_2019_CVPR}.
Bachmann et al. applied masked auto-encoding (MAE) on aligned RGB images, depth maps, and semantic segmentation and learned a shared representation across three domains \cite{bachmann2022multimae}. 
However, although cross-modality representation learning shares a lot of commonalities with heterogeneous tactile representation learning, one significant difference is that heterogeneous tactile representation learning lacks \textit{aligned} data collected with different sensors for different tasks.
It is often ill-defined to infer the resemblance of two tactile images collected from different hardware, which is a prerequisite for many self-supervised learning techniques.
It is therefore desirable to devise a new architecture that is able to learn from \textbf{unaligned} tactile data and produce a representation that is transferable across different tactile sensors and different tasks.

\section{The \texttt{FoTa} Dataset}
\label{sec:dataset}

\begin{figure}[!t]
\centering
\includegraphics[width=\linewidth]{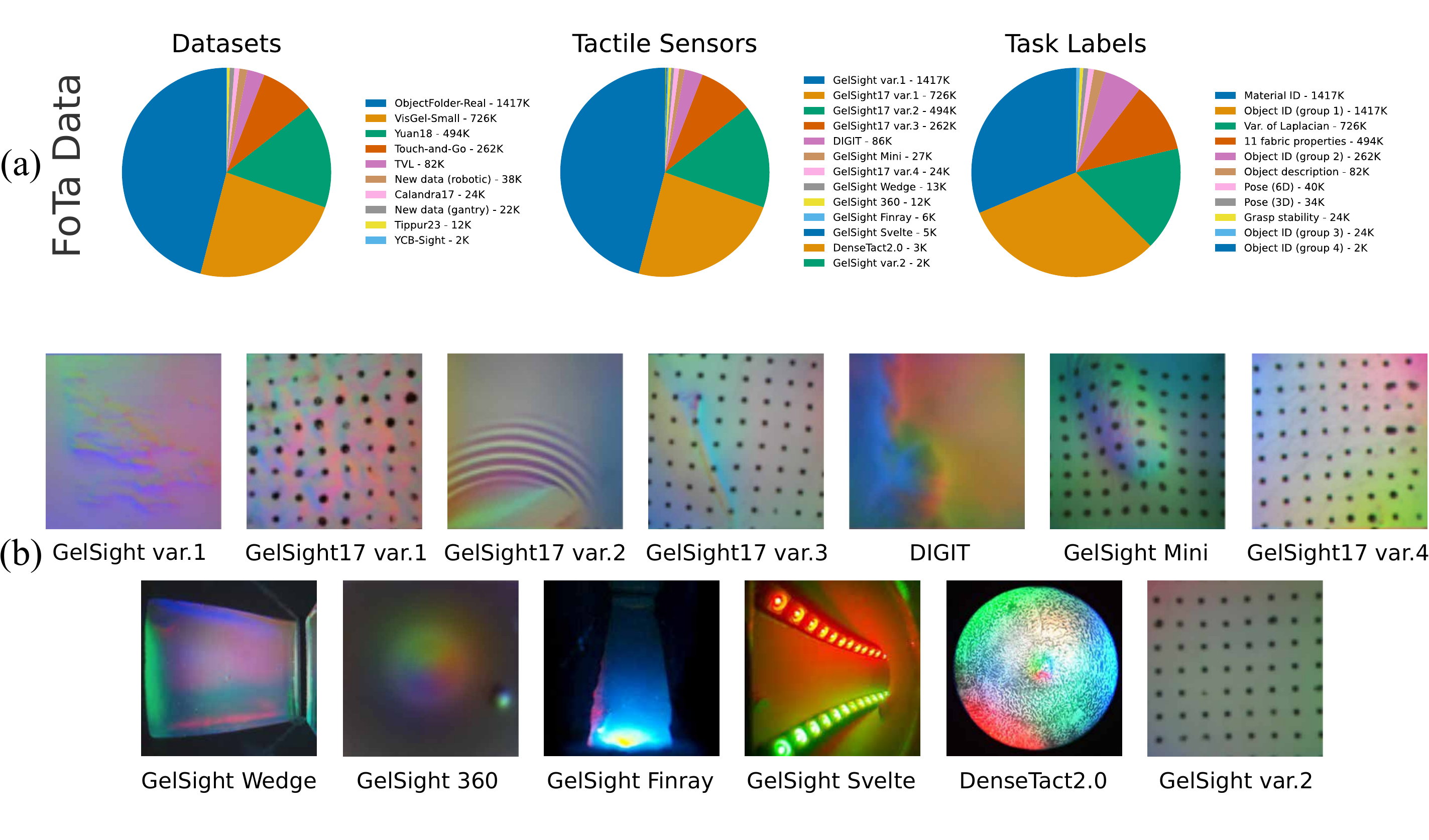}
\caption{\textbf{\texttt{FoTa} dataset visualizations.} (a) We show the mixture and distribution of constituent datasets, sensors, and tasks of the \texttt{FoTa} dataset. Note that not all tasks are utilized in the training of \texttt{T3}. 
(b) We visualize one tactile image from each constituent sensor in \texttt{FoTa}. 
Note that in the training of \texttt{T3}, similar sensors share encoders. For example, \textit{GelSight17 var. \{1-4\}} share the same encoder, and \textit{GelSight var. \{1-2\}} share the same encoder.}
\label{fig:stages}
\end{figure}

We provide a large tactile dataset by aggregating existing public datasets as well as adding new data collected in-house.
The \texttt{FoTa} dataset is provided in a unified, I/O efficient WebDataset~\cite{wds} format.
It is by far the largest and most diverse dataset for tactile perception, with $3,083,452$ data points collected on 13 tactile sensors, and high-quality labels are provided for 11 tasks.
Mixture of constituents of \texttt{FoTa} is illustrated in Fig.~\ref{fig:stages}.a, and example tactile images from each sensor are shown in Fig.~\ref{fig:stages}.b.
Details and statistics about public datasets aggregated in \texttt{FoTa} as well as the pre-processing method are provided in Appendix~\ref{apx:public_datasets}.

Besides aggregating publicly available tactile datasets, we add new data collected with several recently emerged tactile sensors including GelSight Finray~\cite{liu2022gelsight}, GelSight Svelte~\cite{zhao2023gelsight}, GelSight Wedge~\cite{wang2021gelsight}, DenseTact 2.0~\cite{do2022densetact}, and GelSight 360~\cite{tippur2023gelsight360}.
Those new data were also used for final performance benchmarking as well as transferability testing for the proposed learning framework.
Two platforms were built to collect these additional data.

\textbf{A 7-DoF robotic platform} collects tactile data by attaching 2 tactile sensors to a parallel jaw gripper of a 7-DoF Franka Emika Panda robot arm and commanding the robot arm to explore an object that is fixed at a known location.
Robot pose in $SE(3)$ is recorded for each step and then transformed into the sensor's coordinate frame.
Object IDs as well as names are also recorded.

\textbf{A 3-DoF gantry platform} is built with a desktop 3-axis CNC mill.
Six 3D printed probes with different textures are attached to a force/torque sensor, which is then attached to the Z axis of the CNC.
During data collection, a sensor is fixed at a known location, and the CNC machine probes the entire sensing surface of the sensor.
Translational poses, probing forces, and probe IDs are recorded for each tactile image frame.

The two systems are introduced in more detail in Appendix~\ref{apx:datacollection}.

\section{Heterogeneous Tactile Learning with \texttt{T3}}

We propose Transferable Tactile Transformers (\texttt{T3}) as the backbone network to learn from diverse tactile sensors and produce outputs for diverse downstream tasks.

\paragraph{Network architecture}
A training dataset contains inputs from $N_s$ different sensors and $N_t$ distinguished tasks.
\texttt{T3} is constructed with $N_s$ encoders $\{Enc_1, ..., Enc_{N_s}\}$, one for each individual tactile sensor; $N_t$ decoders $\{Dec_1, ..., Dec_{N_t}\}$, each responsible for one downstream task
; and one shared trunk, denoted as $Trunk$.
During both training and inference, we ensure that one batch of data always comes from the same sensor-task pairing.
For a given pair of data point $(X_i, Y_j)$ that is collected from sensor $i$ for task $j$, the corresponding encoder $Enc_i$ and decoder $Dec_j$ are attached to the trunk.
The constituent network components of \texttt{T3} can take different architectures. 
In practice, we use ViT~\cite{dosovitskiy2020image} in the trunk and encoders for its proven learning capacity and scalability in the natural image domain.
For decoders, we use ViT for generative tasks (reconstruction for MAE), ResNet+MLP for pose estimation tasks, and MLP for classification tasks.
More details about network configurations are discussed in Appendix~\ref{apx:network}.

\paragraph{Learning objectives} 
For each task $j\in[1, N_t]$, we define a loss function $L_j$.
Different tasks require different numbers of tactile images as inputs.
For example, tasks like object classification and material classification only require one tactile image to perform, while the pose estimation task requires two tactile images, e.g. $X^1,X^2$, because the prediction goal is the relative pose between the two instances.
We pass each individual tactile image through the encoder and trunk separately, and we then concatenate them together before passing them through the decoder.
The loss between a data pair $(X_i, Y_j)$ or $([X_i^1, X_i^2], Y_j)$ is calculated as
\begin{equation}
\label{eqn:loss_func}
\begin{split}
&loss(X_i, Y_j) = L_j(Y_j, Dec_j(Trunk(Enc_i(X_i)))) \\
&loss([X_i^1, X_i^2], Y_j) = L_j(Y_j, Dec_j(Trunk(Enc_i(X_i^1))\oplus Trunk(Enc_i(X_i^2))))
\end{split}
\end{equation}
where $\oplus$ denotes concatenation.

The training process of \texttt{T3} is split into two pre-training phases and one optional fine-tuning phase.
Pre-training is divided into two phases to improve data utilization and to attain both local fine-grained understanding and global semantic understanding.
Phase 1 focuses on more pixel-level understanding using self-supervised learning, and it is able to utilize all data from the \texttt{FoTa} dataset.
Phase 2 focuses on semantic understanding and it requires suitable task labels that only a subset of the \texttt{FoTa} dataset \changes{possesses}.

\paragraph{Pre-training I: Self-supervised learning with Masked Auto Encoding (MAE)} 
MAE was first proposed as a self-supervised pre-training technique for natural images \cite{he2022masked} and it has gained increased popularity in other domains such as audio processing \cite{huang2022amae}, point-cloud processing \cite{pang2022masked}, and multi-modal representation learning \cite{bachmann2022multimae}.
It operates by first randomly masking out a certain portion of the original inputs, then it constructs a decoder to reconstruct the missing portion.
Neural networks that were pre-trained with MAE have been shown to be able to outperform ones that were trained without MAE in various domains.
This pre-training stage is able to utilize all data in \texttt{FoTa}, including unlabeled data (e.g. intermediate tactile images collected during the settling period of a robot grasp in \cite{calandra2017feeling}), or data with semantic labels for which loss functions are hard to define (e.g. natural language description of a scene in \cite{fu2024tvl}).

In this pre-training stage, one decoder $Dec_0$ that is designed to reconstruct partially masked images is shared by all domains.
$Dec_0$ is constructed with $8$ VisionTransformer blocks \cite{dosovitskiy2020image}.
The loss $L_0$ is L2 pixel-wise loss normalized within each patch, following \cite{he2022masked}.

\paragraph{Pre-training II: Supervised learning with labels distilled from public datasets}
In this stage, \texttt{T3} is trained under supervision from selected task labels in the \texttt{FoTa} dataset.
For each individual task $j\in[1, N_t]$, a decoder $Dec_j$ and a loss function $L_j$ is defined.
We define 10 tasks for this training stage.
\begin{itemize}
    \item Classification tasks include object classification for \cite{calandra2017feeling, yang2022touch}, material classification for \cite{gao2022objectfolder}, and fabric smoothness, fuzziness, textile type classification for \cite{yuan2018active}. Decoders for each of them are defined as MLP neural networks, and each $L_j$ is a cross-entropy loss function.
    \item Regression tasks include $SE(3)$ relative pose estimation bewteen two overlapped tactile images for \cite{wang2021gelsight, lambeta2020digit, gsmini} (with ResNet+MLP decoders), and variance of Laplacian (a metric describing the amount of information contained in tactile images, more details in Appendix~\ref{apx:public_datasets}) estimation for \cite{Li_2019_CVPR} (with MLP decoders). 
    Each loss $L_j$ is a mean squared error loss function.
\end{itemize}

\paragraph{Fine-tuning: Supervised learning with task-specific data}
In this phase, \texttt{T3} is further fine-tuned with data collected for the specific downstream task with the specific sensor that the user aims to use.
This stage can be optional if the sensor-task pairing already exists in the pre-training dataset.
Architecture-wise, \texttt{T3} is configured the same way as it is in Pre-training II but with only one target task and sensor.

\section{Experiments and Discussions}
\label{sec:exp}

In this section we discuss the evaluation performance of \texttt{T3} with numerous ablations and benchmarks.
For both pre-training I and II, we use all data from the \texttt{FoTa} dataset except data collected on the gantry system in-house, including the object classification task and the 3D pose estimation task. Those data are therefore not included for pre-training the foundation model and are set aside for evaluation and fine-tuning only.
We aim to answer the following questions in this section:
\begin{itemize}
    \item How to train efficiently and how much does pre-training improve performance? Sec.~\ref{sec:exp_1}.
    \item Is pre-trained \texttt{T3} \textit{zero-shot} transferable to new tasks and / or sensors? Sec.~\ref{sec:exp_2}.
    \item Does \texttt{FoTa} provide meaningful improvements in longer-horizon robotic tasks? Sec.~\ref{sec:exp_3}.
\end{itemize}

\begin{figure}[!t]
\centering
\includegraphics[width=\linewidth]{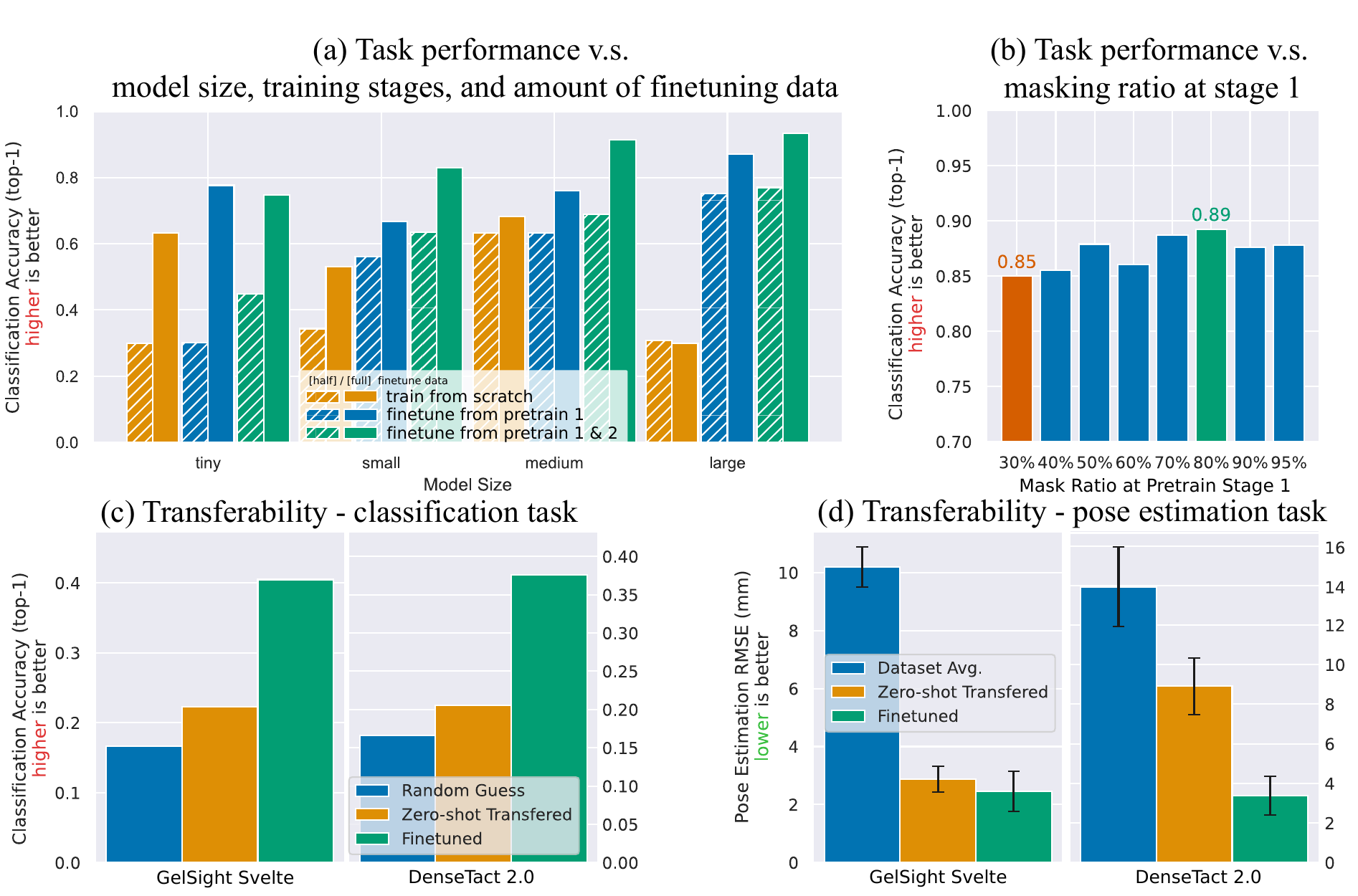}
\caption{\textbf{Experiments on Task Performance and Transferability of \texttt{T3}.} 
(a) Eval performance with 4 network sizes (\texttt{tiny}, \texttt{small}, \texttt{medium}, \texttt{large}), 3 training schemes (train from scratch, fine-tune from pre-train 1, fine-tune from pre-train 2), and 2 amounts of fine-tuning data (half data and full data)
(b) Eval performance with different masking ratios during Pre-training I.
(c) Transferability test on a classification task.
(d) Transferability test on a pose estimation task.}
\label{fig:all_results}
\end{figure}

\subsection{The worthiness of pre-training}
\label{sec:exp_1}
The fine-tuning and evaluation task for this section is an object classification task (6 categories) for 2 sensors (GelSight Wedge, GelSight Mini) with around 3,300 data points for each sensor.
Averaged top-1 classification accuracy from the two sensors is reported. 

\paragraph{The effect of pre-training and the scaling behavior on different network sizes}
We conducted the classification experiment with all combinations of the following: 
(a) 4 network variations: \textit{tiny}-12M, \textit{small}-45M, \textit{medium}-174M, and \textit{large}-308M; 
(b) 3 training schemes: train from scratch, fine-tune from pre-training I, and fine-tune from pre-training I \& II; 
(c) 2 amounts of training data: half data (approx. 1,650 training data) and full data (approx. 3,300 training data). 
All 24 experiments were tested on the same validation dataset, which was separated from the training set. 
The evaluation accuracy is shown in Fig.~\ref{fig:all_results}.a. 
We observed that: 
(a) Pre-training significantly improved the evaluation performance with all network configurations with an median improvement of 24\%. 
(b) The evaluation performance improved with larger networks, where \textit{large}'s classification accuracy was 19\% higher than that of \textit{tiny}. 
However, the performance difference between \textit{medium} and \textit{large} was insignificant. 
(c) When fine-tuned with half of the fine-tuning data, pre-trained networks performed better than trained from scratch networks, and larger pre-trained models performed better than smaller pre-trained models. 
In fact, for \textit{medium} and \textit{large}, the performances between fine-tuned with half data and fine-tuned with full data were very close, signaling those models generalized better thus they require less data to fine-tune to a novel task.

More qualitatively analysis, including visualizations of the encoder attention maps and the trunk attention maps, can be found in Appendix.~\ref{apx:attn}.


\paragraph{Masking ratios in Pre-training Stage I}
We tested 8 different masking ratios in the MAE reconstruction pre-training stage, and we performed the same Pre-training II stage as well as the fine-tuning stage on the 8 tests.
The evaluation results are reported in Fig.~\ref{fig:all_results}.b.
We observe a small performance variance, ranging from 85\% ($mr=30\%$) to 89\% ($mr=80\%$), and a masking ratio of 80\% yielded the highest performance.

\subsection{Zero-shot transferability of pre-trained \texttt{T3} to new sensors or tasks}
\label{sec:exp_2}
We tested to what extent a pre-trained \texttt{T3} can be transferred to a new sensor or downstream task \textit{with} and \textit{without} fine-tuning.
We picked two novel sensors (GelSight Svelte, DensetTact2.0) and two seen tasks (object classification, pose estimation between two overlapped tactile images).
The two seen tasks were pre-trained with data from GelSight Wedge, GelSight Finray, and GelSight Mini.
During this experiment, we use the encoder pre-trained for GelSight Wedge as the encoder for GelSight Svelte, and GelSight Mini's encoder in place of DenseTact2.0's encoder.
Results on the 4 combinations are shown in Fig.~\ref{fig:all_results}.c and Fig.~\ref{fig:all_results}.d.
Zero-shot transfer yielded only minor improvements over random guesses in the object classification task; however, it demonstrated significant improvement over dataset averages in the pose estimation task.
We further fine-tuned the networks with a small amount of 2,000 data points.
The classification accuracy immediately bounced up to 17\% with fine-tuning, and the pose estimation root mean squared error (RMSE) was reduced by 5.5mm for DenseTact2.0.
On the GelSight Svelte sensor, the pose estimation errors before and after fine-tuning were nearly identical and close to optimal.
These results indicate that zero-transferability can be achieved in certain sensor-task pairings, and that better outcomes can be attained with minimal fine-tuning.

\subsection{\texttt{T3} in long-horizon manipulation tasks}
\label{sec:exp_3}

Tactile sensing can be especially helpful in contact-rich manipulation tasks. 
Numerous human studies have shown that the sense of touch is important for humans to perform delicate manipulation tasks such as surgery \cite{eltaib2003tactile, schostek2006artificial}.
However, most state-of-the-art tactile manipulation research opted to train a tactile encoder from scratch, with fresh data collected with the specific sensor-task pairing, due to the lack of a pre-trained tactile encoder.
In the case of long-horizon multi-modal manipulation tasks, training from scratch is often inefficient, due to the sparseness of trajectory rewards and the number of components that need to be trained (e.g. modality encoders and policy networks) especially in many reinforcement learning and behavior cloning applications such as \cite{dong2021tactilerl, wang2024poco}

To investigate whether a pre-trained \texttt{T3} helps to bridge this gap, we designed a robotic precision insertion task with behavior cloning.
The goal of this task is to insert 3 electronics parts: a 3-pin toggle switch, a 12-pin double-stack USB port, and a 17-pin VGA connector, onto a PCB with corresponding mounting holes to each component.
This task requires high precision, where the clearance between the holes on the PCB and the pins on the parts is only $0.4mm$.
Achieving this precision requires active exploration with tactile feedback.
In real-world applications, relying on vision alone is often insufficient due to heavy occlusion.

\begin{figure}[!t]
\centering
\includegraphics[width=\linewidth]{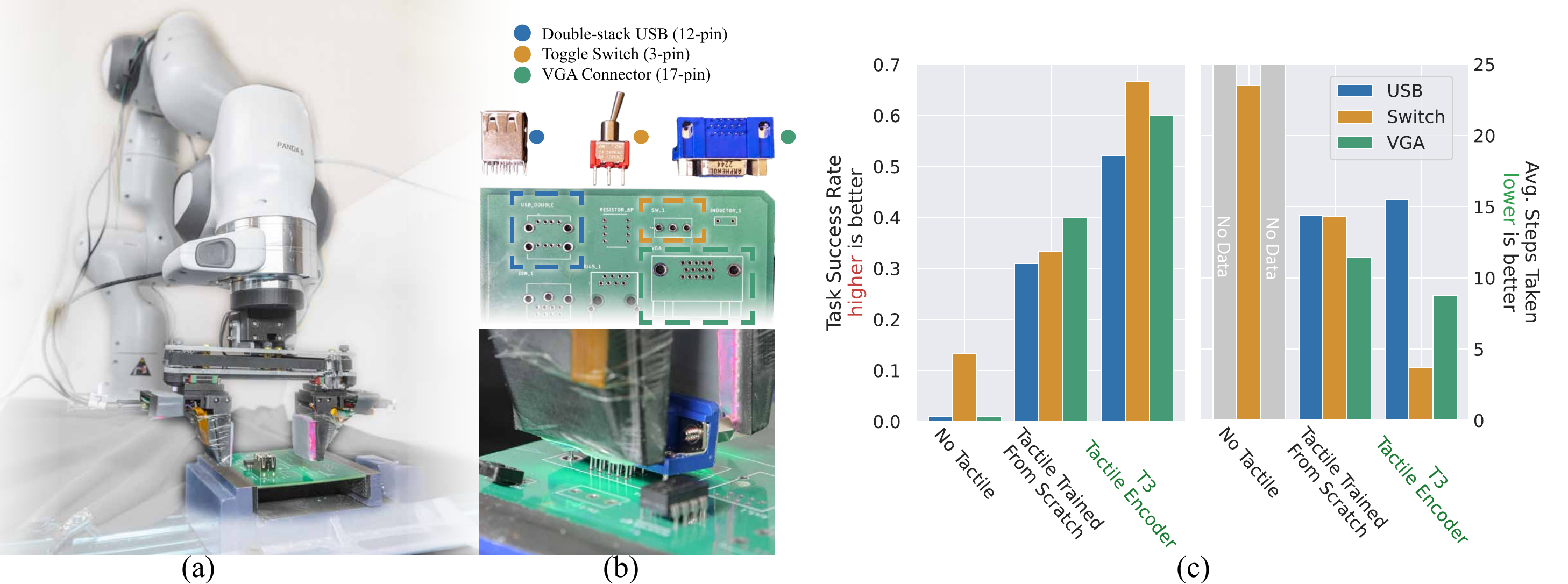}
\caption{\textbf{Real world sub-millimeter robotic insertion tasks.}
(a) Hardware setup.
(b) The 3 evaluation electronics parts and their respective mounting places on a PCB.
(c) Insertion success rate and average steps taken for each task. Policies trained with the \texttt{T3} tactile encoder achieved the highest success rate and lowest averaged episode length.
}
\label{fig:insertion}
\end{figure}

Our experiment setup consists of two GelSight Wedge tactile sensors mounted on a parallel jaw gripper, which is attached to the end-effector of a 7-DoF Franka Emika Panda robot, a PCB board fixed at a known location on the workbench, the 3 electronics parts, and a RGB camera on the side of the workbench.
An illustration of the setup is shown in Fig.~\ref{fig:insertion} a and the parts are shown in Fig.~\ref{fig:insertion}.b.
Data is collected by controlling the robot in "guide mode", i.e. under zero stiffness control, and a human operator manually grabs and moves the end-effector until the part is fully inserted.
At both data collection and inference time, the sockets are fixed, and we add randomizations to the part-in-hand pose and gripper starting pose before each insertion, both generated from a 3D zero-mean, $1mm$ standard deviation normal distribution. 

We train and evaluate 3 policies: a baseline policy without tactile input, a policy with tactile input encoded by a neural network trained from scratch, and a policy with tactile input encoded by \texttt{T3}.
Besides tactile inputs, all 3 policies have access to the same robot state modality encoded by a MLP, and external vision modality encoded by a pre-trained ResNet18.
All three policies take observations of the current step as inputs and predict a 3-DoF action which the robot executes at the next time step.
At inference time, the robot executes predicted 3-DoF actions in about 2Hz for up to 30 steps.
An episode is deemed successful if the robot successfully inserts the component within 30 steps.
The success rate and the average steps taken of successful episodes across 15 episodes are reported in Fig.~\ref{fig:insertion}.c.

The results show that (1) the tactile modality is vital for this electronics insertion task, where the vision-only policy failed all tests to insert the two more challenging parts; (2) using a pre-trained \texttt{T3} as the tactile encoder for this policy helped further improve the overall performance, where the task success rate is higher across all three parts; (3) \texttt{T3} also helped to reduce the number of tactile exploration steps needed to insert a part.


\section{Limitations and Future Works}
\label{sec:limitations}
\vspace{-0.5em}
One limitation of the current training pipeline is that the \texttt{FoTa} dataset is unbalanced.
Data collected with the 2 most popular sensors constitutes over 50\% of the entire dataset.
Therefore, the trained policy could also be biased towards those more popular sensors.
Another limitation is that the current architecture of \texttt{T3} focuses on per-image encoding and \texttt{T3} is trained with explicit labels during pre-training II and fine-tuning.
Representation learning on tactile image sequences with sparse or implicit labels can be a future direction of this work. 
\changes{Furthermore, \texttt{T3} is currently limited to learning representations for camera-based tactile sensors. 
Extending the \texttt{FoTa} dataset and the \texttt{T3} framwork to non-camera-based sensors is another future line of work.}

\section{Conclusions}
\label{sec:conclusion}
\vspace{-0.5em}
This paper presents \texttt{T3}, a large camera-based tactile sensing framework that transfers across multiple sensors and tasks; and \texttt{FoTa}, a tactile dataset that is by far the largest in quantity and most diverse in sensors and downstream tasks.
Experiments show that \texttt{T3} pre-trained with \texttt{FoTa} improves performance for various tasks, including longer-horizon manipulation tasks such as sub-millimeter electronics insertion, and that \texttt{T3} transfers to a new sensor-task pairing with little fine-tuning.


\clearpage
\acknowledgments{The authors thank Kaiming He for the many insights and discussions around network architecture and training. This paper is partially funded by Toyota Research Institute and Amazon Science Hub. The authors also thank Won Kyung Do from the Stanford ARMLab for providing a DenseTact2.0 for the experiments conducted in the paper.}


\bibliographystyle{IEEEtran}
\bibliography{refs}

\begin{thebibliography}{10}
\providecommand{\url}[1]{#1}
\csname url@rmstyle\endcsname
\providecommand{\newblock}{\relax}
\providecommand{\bibinfo}[2]{#2}
\providecommand\BIBentrySTDinterwordspacing{\spaceskip=0pt\relax}
\providecommand\BIBentryALTinterwordstretchfactor{4}
\providecommand\BIBentryALTinterwordspacing{\spaceskip=\fontdimen2\font plus
\BIBentryALTinterwordstretchfactor\fontdimen3\font minus \fontdimen4\font\relax}
\providecommand\BIBforeignlanguage[2]{{%
\expandafter\ifx\csname l@#1\endcsname\relax
\typeout{** WARNING: IEEEtran.bst: No hyphenation pattern has been}%
\typeout{** loaded for the language `#1'. Using the pattern for}%
\typeout{** the default language instead.}%
\else
\language=\csname l@#1\endcsname
\fi
#2}}

\bibitem{calandra2018more}
R.~Calandra, A.~Owens, D.~Jayaraman, J.~Lin, W.~Yuan, J.~Malik, E.~H. Adelson, and S.~Levine, ``More than a feeling: Learning to grasp and regrasp using vision and touch,'' \emph{IEEE Robotics and Automation Letters}, vol.~3, no.~4, pp. 3300--3307, 2018.

\bibitem{calandra2017feeling}
R.~Calandra, A.~Owens, M.~Upadhyaya, W.~Yuan, J.~Lin, E.~H. Adelson, and S.~Levine, ``The feeling of success: Does touch sensing help predict grasp outcomes?'' \emph{arXiv preprint arXiv:1710.05512}, 2017.

\bibitem{sunil2023visuotactile}
N.~Sunil, S.~Wang, Y.~She, E.~Adelson, and A.~R. Garcia, ``Visuotactile affordances for cloth manipulation with local control,'' in \emph{Conference on Robot Learning}.\hskip 1em plus 0.5em minus 0.4em\relax PMLR, 2023, pp. 1596--1606.

\bibitem{yuan2018active}
W.~Yuan, Y.~Mo, S.~Wang, and E.~Adelson, ``Active clothing material perception using tactile sensing and deep learning,'' 2018.

\bibitem{liu2022gelsight}
S.~Q. Liu and E.~H. Adelson, ``Gelsight fin ray: Incorporating tactile sensing into a soft compliant robotic gripper,'' in \emph{2022 IEEE 5th International Conference on Soft Robotics (RoboSoft)}.\hskip 1em plus 0.5em minus 0.4em\relax IEEE, 2022, pp. 925--931.

\bibitem{girdhar2023imagebind}
R.~Girdhar, A.~El-Nouby, Z.~Liu, M.~Singh, K.~V. Alwala, A.~Joulin, and I.~Misra, ``Imagebind: One embedding space to bind them all,'' in \emph{Proceedings of the IEEE/CVF Conference on Computer Vision and Pattern Recognition}, 2023, pp. 15\,180--15\,190.

\bibitem{bachmann2022multimae}
R.~Bachmann, D.~Mizrahi, A.~Atanov, and A.~Zamir, ``Multimae: Multi-modal multi-task masked autoencoders,'' 2022.

\bibitem{huang2022amae}
P.-Y. Huang, H.~Xu, J.~Li, A.~Baevski, M.~Auli, W.~Galuba, F.~Metze, and C.~Feichtenhofer, ``Masked autoencoders that listen,'' in \emph{NeurIPS}, 2022.

\bibitem{wang2024poco}
L.~Wang, J.~Zhao, Y.~Du, E.~H. Adelson, and R.~Tedrake, ``Poco: Policy composition from and for heterogeneous robot learning,'' \emph{arXiv preprint arXiv:2402.02511}, 2024.

\bibitem{shah2023mutex}
R.~Shah, R.~Mart{\'\i}n-Mart{\'\i}n, and Y.~Zhu, ``Mutex: Learning unified policies from multimodal task specifications,'' \emph{arXiv preprint arXiv:2309.14320}, 2023.

\bibitem{si2024difftactile}
Z.~Si, G.~Zhang, Q.~Ben, B.~Romero, Z.~Xian, C.~Liu, and C.~Gan, ``Difftactile: A physics-based differentiable tactile simulator for contact-rich robotic manipulation,'' \emph{arXiv preprint arXiv:2403.08716}, 2024.

\bibitem{gelsight2017}
\BIBentryALTinterwordspacing
S.~Dong, W.~Yuan, and E.~H. Adelson, ``Improved gelsight tactile sensor for measuring geometry and slip,'' in \emph{2017 IEEE/RSJ International Conference on Intelligent Robots and Systems (IROS)}.\hskip 1em plus 0.5em minus 0.4em\relax IEEE, Sept. 2017. [Online]. Available: \url{http://dx.doi.org/10.1109/IROS.2017.8202149}
\BIBentrySTDinterwordspacing

\bibitem{wang2021gelsight}
S.~Wang, Y.~She, B.~Romero, and E.~Adelson, ``Gelsight wedge: Measuring high-resolution 3d contact geometry with a compact robot finger,'' in \emph{2021 IEEE International Conference on Robotics and Automation (ICRA)}.\hskip 1em plus 0.5em minus 0.4em\relax IEEE, 2021, pp. 6468--6475.

\bibitem{do2022densetact}
W.~K. Do and M.~Kennedy, ``Densetact: Optical tactile sensor for dense shape reconstruction,'' in \emph{2022 International Conference on Robotics and Automation (ICRA)}.\hskip 1em plus 0.5em minus 0.4em\relax IEEE, 2022, pp. 6188--6194.

\bibitem{li2018slip}
J.~Li, S.~Dong, and E.~Adelson, ``Slip detection with combined tactile and visual information,'' in \emph{2018 IEEE International Conference on Robotics and Automation (ICRA)}.\hskip 1em plus 0.5em minus 0.4em\relax IEEE, 2018, pp. 7772--7777.

\bibitem{yuan2015measurement}
W.~Yuan, R.~Li, M.~A. Srinivasan, and E.~H. Adelson, ``Measurement of shear and slip with a gelsight tactile sensor,'' in \emph{2015 IEEE International Conference on Robotics and Automation (ICRA)}.\hskip 1em plus 0.5em minus 0.4em\relax IEEE, 2015, pp. 304--311.

\bibitem{yamaguchi2016combining}
A.~Yamaguchi and C.~G. Atkeson, ``Combining finger vision and optical tactile sensing: Reducing and handling errors while cutting vegetables,'' in \emph{2016 IEEE-RAS 16th International Conference on Humanoid Robots (Humanoids)}.\hskip 1em plus 0.5em minus 0.4em\relax IEEE, 2016, pp. 1045--1051.

\bibitem{tippur2023gelsight360}
M.~H. Tippur and E.~H. Adelson, ``Gelsight360: An omnidirectional camera-based tactile sensor for dexterous robotic manipulation,'' in \emph{2023 IEEE International Conference on Soft Robotics (RoboSoft)}.\hskip 1em plus 0.5em minus 0.4em\relax IEEE, 2023, pp. 1--8.

\bibitem{zhao2023gelsight}
J.~Zhao and E.~H. Adelson, ``Gelsight svelte: A human finger-shaped single-camera tactile robot finger with large sensing coverage and proprioceptive sensing,'' in \emph{2023 IEEE/RSJ International Conference on Intelligent Robots and Systems (IROS)}.\hskip 1em plus 0.5em minus 0.4em\relax IEEE, 2023, pp. 8979--8984.

\bibitem{sun2022soft}
H.~Sun, K.~J. Kuchenbecker, and G.~Martius, ``A soft thumb-sized vision-based sensor with accurate all-round force perception,'' \emph{Nature Machine Intelligence}, vol.~4, no.~2, 2022.

\bibitem{liu2023gelsight}
S.~Q. Liu, Y.~Ma, and E.~H. Adelson, ``Gelsight baby fin ray: A compact, compliant, flexible finger with high-resolution tactile sensing,'' in \emph{2023 IEEE International Conference on Soft Robotics (RoboSoft)}.\hskip 1em plus 0.5em minus 0.4em\relax IEEE, 2023, pp. 1--8.

\bibitem{ma2024scalable}
Y.~Ma, A.~Agarwal, S.~Q. Liu, W.~Yuan, and E.~H. Adelson, ``Scalable simulation-guided compliant tactile finger design,'' in \emph{2024 IEEE 7th International Conference on Soft Robotics (RoboSoft)}.\hskip 1em plus 0.5em minus 0.4em\relax IEEE, 2024, pp. 1068--1074.

\bibitem{ma2024gellink}
Y.~Ma, J.~Zhao, and E.~Adelson, ``Gellink: A compact multi-phalanx finger with vision-based tactile sensing and proprioception,'' \emph{arXiv preprint arXiv:2403.14887}, 2024.

\bibitem{zhao2023novel}
F.~Zhao, B.~Huang, M.~Li, M.~Li, Z.~Fu, Z.~Lei, and M.~Li, ``A novel tactile palm for robotic object manipulation,'' 2023.

\bibitem{jiang2021vision}
J.~Jiang, G.~Cao, D.~F. Gomes, and S.~Luo, ``Vision-guided active tactile perception for crack detection and reconstruction,'' in \emph{2021 29th Mediterranean Conference on Control and Automation (MED)}.\hskip 1em plus 0.5em minus 0.4em\relax IEEE, 2021, pp. 930--936.

\bibitem{Li_2019_CVPR}
Y.~Li, J.-Y. Zhu, R.~Tedrake, and A.~Torralba, ``Connecting touch and vision via cross-modal prediction,'' in \emph{The IEEE Conference on Computer Vision and Pattern Recognition (CVPR)}, June 2019.

\bibitem{fu2024tvl}
L.~Fu, G.~Datta, H.~Huang, W.~C.-H. Panitch, J.~Drake, J.~Ortiz, M.~Mukadam, M.~Lambeta, R.~Calandra, and K.~Goldberg, ``A touch, vision, and language dataset for multimodal alignment,'' \emph{arXiv preprint arXiv:2402.13232}, 2024.

\bibitem{suresh2021efficient}
S.~Suresh, Z.~Si, J.~G. Mangelson, W.~Yuan, and M.~Kaess, ``Efficient shape mapping through dense touch and vision,'' \emph{arXiv preprint arXiv:2109.09884}, 2021.

\bibitem{yang2022touch}
F.~Yang, C.~Ma, J.~Zhang, J.~Zhu, W.~Yuan, and A.~Owens, ``Touch and go: Learning from human-collected vision and touch,'' \emph{arXiv preprint arXiv:2211.12498}, 2022.

\bibitem{gao2022objectfolder}
R.~Gao, Z.~Si, Y.-Y. Chang, S.~Clarke, J.~Bohg, L.~Fei-Fei, W.~Yuan, and J.~Wu, ``Objectfolder 2.0: A multisensory object dataset for sim2real transfer,'' in \emph{Proceedings of the IEEE/CVF conference on computer vision and pattern recognition}, 2022, pp. 10\,598--10\,608.

\bibitem{deng2009imagenet}
J.~Deng, W.~Dong, R.~Socher, L.-J. Li, K.~Li, and L.~Fei-Fei, ``Imagenet: A large-scale hierarchical image database,'' in \emph{2009 IEEE conference on computer vision and pattern recognition}.\hskip 1em plus 0.5em minus 0.4em\relax Ieee, 2009, pp. 248--255.

\bibitem{he2016deep}
K.~He, X.~Zhang, S.~Ren, and J.~Sun, ``Deep residual learning for image recognition,'' in \emph{Proceedings of the IEEE conference on computer vision and pattern recognition}, 2016, pp. 770--778.

\bibitem{radford2021learning}
A.~Radford, J.~W. Kim, C.~Hallacy, A.~Ramesh, G.~Goh, S.~Agarwal, G.~Sastry, A.~Askell, P.~Mishkin, J.~Clark, \emph{et~al.}, ``Learning transferable visual models from natural language supervision,'' in \emph{International conference on machine learning}.\hskip 1em plus 0.5em minus 0.4em\relax PMLR, 2021, pp. 8748--8763.

\bibitem{li2022see}
H.~Li, Y.~Zhang, J.~Zhu, S.~Wang, M.~A. Lee, H.~Xu, E.~Adelson, L.~Fei-Fei, R.~Gao, and J.~Wu, ``See, hear, and feel: Smart sensory fusion for robotic manipulation,'' \emph{arXiv preprint arXiv:2212.03858}, 2022.

\bibitem{qi2023general}
H.~Qi, B.~Yi, S.~Suresh, M.~Lambeta, Y.~Ma, R.~Calandra, and J.~Malik, ``General in-hand object rotation with vision and touch,'' in \emph{Conference on Robot Learning}.\hskip 1em plus 0.5em minus 0.4em\relax PMLR, 2023, pp. 2549--2564.

\bibitem{wang2024hpt}
\BIBentryALTinterwordspacing
L.~Wang, X.~Chen, J.~Zhao, and K.~He, ``Scaling proprioceptive-visual learning with heterogeneous pre-trained transformers,'' \emph{arXiv}, 2024. [Online]. Available: \url{https://arxiv.org/abs/2409.20537}
\BIBentrySTDinterwordspacing

\bibitem{zhao2023fingerslam}
J.~Zhao, M.~Bauza, and E.~H. Adelson, ``Fingerslam: Closed-loop unknown object localization and reconstruction from visuo-tactile feedback,'' in \emph{2023 IEEE International Conference on Robotics and Automation (ICRA)}.\hskip 1em plus 0.5em minus 0.4em\relax IEEE, 2023, pp. 8033--8039.

\bibitem{dave2024multimodal}
V.~Dave, F.~Lygerakis, and E.~Rueckert, ``Multimodal visual-tactile representation learning through self-supervised contrastive pre-training,'' \emph{arXiv preprint arXiv:2401.12024}, 2024.

\bibitem{chen2022visuo}
Y.~Chen, M.~Van~der Merwe, A.~Sipos, and N.~Fazeli, ``Visuo-tactile transformers for manipulation,'' in \emph{6th Annual Conference on Robot Learning}, 2022.

\bibitem{wds}
T.~W. developer team, ``Webdataset,'' \url{https://github.com/webdataset/webdataset}, 2020, [Online].

\bibitem{dosovitskiy2020image}
A.~Dosovitskiy, L.~Beyer, A.~Kolesnikov, D.~Weissenborn, X.~Zhai, T.~Unterthiner, M.~Dehghani, M.~Minderer, G.~Heigold, S.~Gelly, \emph{et~al.}, ``An image is worth 16x16 words: Transformers for image recognition at scale,'' \emph{arXiv preprint arXiv:2010.11929}, 2020.

\bibitem{he2022masked}
K.~He, X.~Chen, S.~Xie, Y.~Li, P.~Doll{\'a}r, and R.~Girshick, ``Masked autoencoders are scalable vision learners,'' in \emph{Proceedings of the IEEE/CVF conference on computer vision and pattern recognition}, 2022, pp. 16\,000--16\,009.

\bibitem{pang2022masked}
Y.~Pang, W.~Wang, F.~E. Tay, W.~Liu, Y.~Tian, and L.~Yuan, ``Masked autoencoders for point cloud self-supervised learning,'' in \emph{European conference on computer vision}.\hskip 1em plus 0.5em minus 0.4em\relax Springer, 2022, pp. 604--621.

\bibitem{lambeta2020digit}
M.~Lambeta, P.-W. Chou, S.~Tian, B.~Yang, B.~Maloon, V.~R. Most, D.~Stroud, R.~Santos, A.~Byagowi, G.~Kammerer, \emph{et~al.}, ``Digit: A novel design for a low-cost compact high-resolution tactile sensor with application to in-hand manipulation,'' \emph{IEEE Robotics and Automation Letters}, vol.~5, no.~3, pp. 3838--3845, 2020.

\bibitem{gsmini}
G.~Inc., ``{GelSight Mini},'' \url{https://www.gelsight.com/gelsightmini/}, 2022, [Online].

\bibitem{eltaib2003tactile}
M.~Eltaib and J.~Hewit, ``Tactile sensing technology for minimal access surgery----a review,'' \emph{Mechatronics}, vol.~13, no.~10, pp. 1163--1177, 2003.

\bibitem{schostek2006artificial}
S.~Schostek, C.-N. Ho, D.~Kalanovic, and M.~O. Schurr, ``Artificial tactile sensing in minimally invasive surgery--a new technical approach,'' \emph{Minimally invasive therapy \& allied technologies}, vol.~15, no.~5, pp. 296--304, 2006.

\bibitem{dong2021tactilerl}
S.~Dong, D.~K. Jha, D.~Romeres, S.~Kim, D.~Nikovski, and A.~Rodriguez, ``Tactile-rl for insertion: Generalization to objects of unknown geometry,'' 2021.

\end{thebibliography}

\clearpage

\appendix

\section{Appendix}

\subsection{Visualization of attention maps}
\label{apx:attn}
\begin{figure}[!h]
\centering
\includegraphics[width=\linewidth]{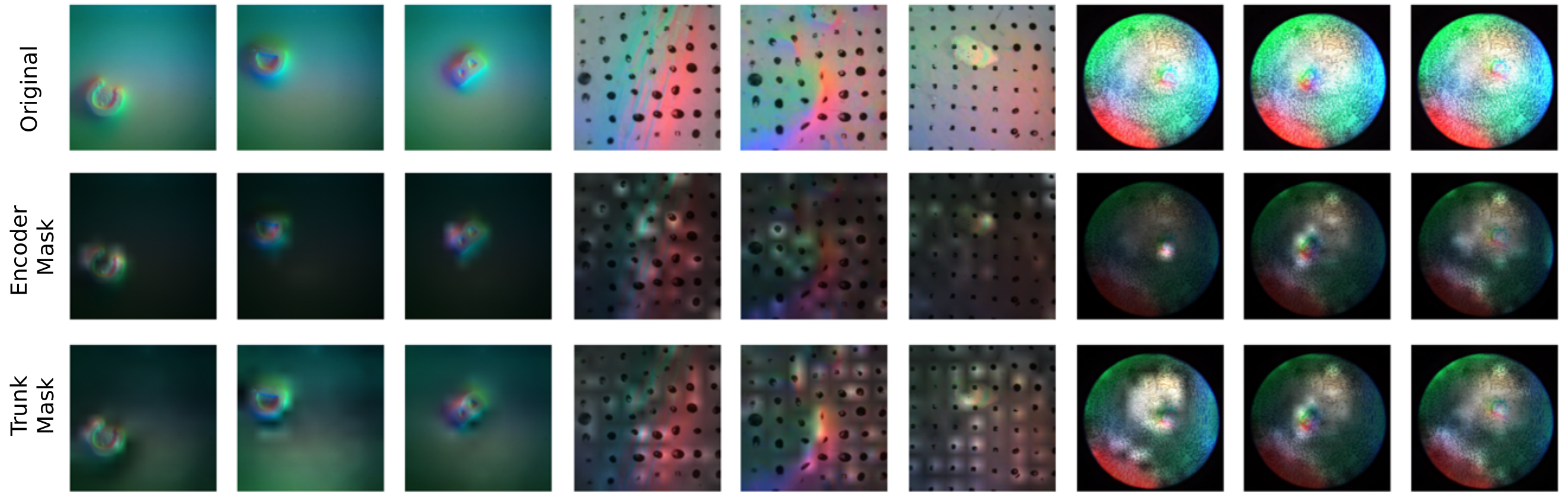}
\caption{\changes{\textbf{Visualization of self-attention weights of the encoders and the trunk of pre-trained \texttt{T3}.}}}
\changes{Top: the original tactile images. 
Middle: encoder attention weights applied on the original images as masks. 
Bottom: trunk attention weights applied on the original images as masks}
\label{fig:attn}
\end{figure}
\changes{
We calculate the attention weights $attn_{i} = softmax(\frac{Q \cdot K}{scale})$ at layer $i$ of each self-attention layers for all encoders and the trunk.
We then obtain a joint attention map, calculated as the product of the attention weights of all layers within each encoder or trunk, $map = \prod_i attn_{i}$.
This joint attention map is then normalized and applied on the original tactile images as masks. 
A few tactile images as well as them applied with encoder masks and trunk masks are visualized in Fig.~\ref{fig:attn}. 
Qualitatively, we observe that the encoder self-attentions highlight the region of the contact object, whereas the trunk self-attentions are more spread out and they cover other features like the markers on a sensor. 
This result seems to show that the encoder is more focused on the sensor agnostic regions, i.e. the "hot" regions of the contact, while the trunk also attends to sensor specific areas.
}

\subsection{Public datasets aggregated in the \texttt{FoTa} dataset}
\label{apx:public_datasets}
The \texttt{FoTa} dataset does not modify the images or labels from the constituent datasets.
Pre-processing was performed on each of them to encode the tactile images into $.jpg$ format pictures and to extract all labels into $.json$ format dictionaries.
In one of the constituent public datasets, VisGel~\cite{Li_2019_CVPR}, more than half of the included tactile images are "flat" images, i.e. tactile images captured when there is no contact.
We sub-sampled VisGel to reduce the amount of flat images.
With an image frame $I_i$ and a true flat frame $I_f$, the variance of Laplacian of the different image $\sigma_i$ is calculated with Eqn.~\ref{eqn:sigma}. 
We then remove all images whose $\sigma < \sigma_{t}$.
We empirically chose $\sigma_{t} = 4.24$ to remove most flat images.
\begin{equation}
\label{eqn:sigma}
    \sigma_i = \sqrt{Var(Laplacian(I_i - I_f))}, i \in \{1...N\} \\
\end{equation}
Note that we do not aim to eliminate flat images entirely as flat images also contain important information such as the sensor geometry and the illumination design.
Statistics of the aggregated datasets in \texttt{FoTa} are listed in Tab.~\ref{tab:dataset}.
 
\subsection{Data format}
\label{apx:dataformat}
Each constituent dataset of \texttt{FoTa} is split into one \texttt{train} dataset and one \texttt{val} dataset, and each of them is packaged in a pre-sharded WebDataset~\cite{wds} format.
Each shard of the \texttt{FoTa} dataset contains up to $10,000$ data points packed in one \textit{.tar} archive, where each data point is composed of one \textit{.jpg} tactile image and one \textit{.json} file for various task labels.
Utility scripts are also provided to add new data to the \texttt{FoTa} dataset or to modify the pre-processing procedures on the constituent public datasets.

\begin{table}[!htb]
  \caption{Network Sizes}
  \label{tab:network_size}
  \centering
  \begin{tabular}{cccccc}
    \toprule
        & $D_{enc}$      & $H_{enc}$ & $L_{enc}$ & $L_{tru}$   & \shortstack[c]{Params\\incl. all encoders and decoders}\\
        \midrule
        \texttt{T3-tiny} & 192 & 3 & 3 & 9 & 12M \\
        \texttt{T3-small} & 384 & 6 & 3 & 9 & 45M \\
        \texttt{T3-medium} & 768 & 12 & 3 & 9 & 174M \\
        \texttt{T3-large} & 1024 & 8 & 3 & 9 & 308M \\
    \bottomrule
  \end{tabular}
\end{table}

\begin{table}[!b]
  \caption{Dataset Statistics}
  \label{tab:dataset}
  \centering
  \begin{tabular}{p{18mm}p{25mm}p{25mm}p{15mm}cp{15mm}}
    \toprule
    Source      & Objects      & Labels & Sensor & Marker   & Size \\
    \midrule
    VisGel\cite{Li_2019_CVPR}-small* & Household objects and toys. Flat images from the original dataset were filtered. & Variance of the Laplacian of diff. images  & GelSight'17 \cite{gelsight2017} & Yes & 726,740 \\
    \midrule
    TVL \cite{fu2024tvl}     & Household objects   & semantic object description    & DIGIT \cite{lambeta2020digit} & No  & 82,463  \\
    \midrule
    Touch-and-Go \cite{yang2022touch} & Natural objects & Object IDs & GelSight'17 \cite{gelsight2017} & Yes & 262,082 \\
    \midrule
    Calandra'17 \cite{calandra2017feeling} & Household objects & Object IDs, grasp outcome & GelSight'17 \cite{gelsight2017} & Yes & 24,118 \\
    \midrule
    Yuan'18 \cite{yuan2018active} & Clothes & 11 properties such as smoothness, softness & GelSight'17 \cite{gelsight2017} & Yes  & 494,655 \\
    \midrule
    YCB-Sight \cite{suresh2021efficient} & Household objects from the YCB object set & Object IDs, 6D poses & GelSight variant No. 1 and\newline simulation & No & 480 real \newline 1,800 sim \\
    \midrule
    ObjectFolder-real \cite{gao2022objectfolder} & Household objects & Object, material IDs & GelSight variant No. 2 & No & 1,417,600 \\
    \midrule
    Tippur'23 \cite{tippur2023gelsight360} & Probing sphere & 3D poses & GelSight 360 & No & 13,341 \\
    \midrule
    \multirow{19}*{\shortstack[l]{New data \\collected\\in-house}} & \multirow{6}*{Tool objects} & \multirow{6}*{\shortstack[l]{Object IDs\\ 6D poses}} & GelSight Wedge \cite{wang2021gelsight} & No &  10,000 \\
        \cmidrule(lr){4-6}
     &  & & GelSight Mini \cite{gsmini} & Yes\&No & 24,000 \\
        \cmidrule(lr){4-6}
     &  & & DIGIT \cite{lambeta2020digit} & No & 4,000 \\
    \cmidrule(lr){2-6}
     & \multirow{12}*{\shortstack[l]{Probes with \\engraved letters}} & \multirow{12}*{\shortstack[l]{Object IDs\\ 3D poses}} & GelSight Wedge \cite{wang2021gelsight} & No &  3,516 \\
        \cmidrule(lr){4-6}
     & & & GelSight Finray \cite{liu2023gelsight} \newline 2 variants & No &  6,432 \\
        \cmidrule(lr){4-6}
     & & & GelSight Svelte \cite{zhao2023gelsight}& No &  5,335 \\
        \cmidrule(lr){4-6}
     & & & DenseTact2.0 \cite{do2022densetact} & Yes &  3,531 \\
     \cmidrule(lr){4-6}
    & & & GelSight Mini \cite{gsmini} & Yes &  3,359 \\
    \bottomrule
    \multicolumn{6}{r}{* VisGel-small is down-sampled from the original VisGel and most flat images are removed.} \\
  \end{tabular}
\end{table}

\subsection{Data collection platforms}
\label{apx:datacollection}

\subsubsection{7-DoF robotic platform}
We built a robotic data collection platform with a 7-DoF Franka Emika robotic arm with a custom-made parallel jaw gripper attached at the end-effector.
Two tactile sensors are attached at the tips of the parallel jaw gripper, and the transformation between the camera frame and the end-effector frame is accurately measured.
During tactile data collection, a test object, which is assumed to be rigid enough to be clamped firmly, is fixed on the workbench of the robot at a known location. 
The robot explores the object with random movements on all 6 dimensions (translation and rotation), and tactile images from both sensors as well as the robot pose are collected at each step.

\subsubsection{3-DoF gantry platform}
Another gantry-based 3-DoF data collection platform was built from a desktop CNC milling machine.
The spindle was replaced with a 3D-printed probe, which is attached to the z-axis of the CNC with a 6-axis force torque sensor (MMS101, Mitsumi Electric Company).
The probe can be easily swapped out, and we designed in total 6 probes with different textures.
During data collection, a tactile sensor is attached firmly to the CNC bed, and the CNC probes the tactile sensor by applying vertical force until the z-axis force reaches a threshold (uniformly sampled from $0.3 - 2.1 N$ for most sensors, except GelSight Mini and GelSight Svelte for which we use $1.5 - 5 N$ due to higher gel stiffness).
Tactile images, probing positions in the sensor's coordinate frame (3D translational poses), and the force values are recorded.

\subsection{Network configurations}
\label{apx:network}
In principle, all components of \texttt{T3} can adopt different architectures, provided they maintain coherence. 
This section lists the specific architecture and hyper-parameter choices used in this work.

\paragraph{Encoders}
We use VisionTransformer (ViT) \cite{dosovitskiy2020image} blocks as the encoders for tactile images.
An image is first resized to $256\times 256$ then center-cropped to $224\times 224$, before split into $196$ image blocks each measures $16\times 16$ pixels.
Augmentation is performed during training.
Specifically, color jittering is performed for all tasks.
Random cropping, horizontal and vertical flipping are performed for all tasks except pose estimation tasks, which rely on the spatial information to make predictions.
We use a MLP ratio of $4.0$, an embedding dimension of $D_{enc}$, $H_{enc}$ heads, and $L_{enc}$ layers across encoders for all sensors.

\paragraph{Trunk}
We use ViT blocks as the trunk.
We use a MLP ratio of $4.0$, an embedding dimension of $D_{tru} = D_{enc}$, $H_{tru} = H_{enc}$ heads, and $L_{tru}$ layers.

\paragraph{Decoders}
The decoder architecture varies based on the tasks.
The reconstruction decoder for masked auto encoding during pre-training I is also a ViT, with a MLP ratio of $4.0$, an embedding dimension of $512$, $16$ heads, and $8$ layers. 
    No pooling is performed after the trunk for this decoder.
    
The decoders for $\sigma_i$ regression (Eqn.~\ref{eqn:sigma}) and all classification tasks are MLP decoders.
    We used the same size of 3 hidden layers $[256, 128, 64]$ for all these MLP decoders.
    The output dimension varies for the classification tasks depending on the number of categories.
    Only the classification token is kept from the output of the trunk before it is passed through these decoders.

The decoders for both 3D and 6D pose estimation tasks are constructed with 2 ResNet \cite{he2016deep} blocks followed by average pooling then a MLP with hidden layers $[256, 64]$.
    Before passing through the token outputs from the trunk into the ResNet blocks, we first remove the classification token, then we reshape the output from $[B, T, C]$ to $[B, C, \sqrt{T-1}, \sqrt{T-1}]$.
    Intuitively, we transform the embedding space back to a 2D map, on which convolution is performed by the ResNet blocks.

Choices of hyper parameters are listed in Tab.~\ref{tab:network_size}.


\end{document}